\documentclass[lettersize,journal]{IEEEtran}
\usepackage{amsmath,amsfonts}
\usepackage{algorithmic}
\usepackage{algorithm}
\usepackage{float}
\usepackage{array}
\usepackage{graphicx}
\usepackage{subcaption}
\usepackage{array} 
\usepackage[caption=false,font=normalsize,labelfont=sf,textfont=sf]{subfig}
\usepackage{textcomp}
\usepackage{stfloats}
\usepackage{url}
\usepackage{verbatim}
\usepackage{graphicx}
\usepackage{cite}
\usepackage{todonotes}
\newcommand{\mycomment}[1]{}

\begin{document}

\title{Pareto Data Framework: Steps Towards
Resource-Efficient Decision Making Using Minimum
Viable Data (MVD)}

\author{Tashfain Ahmed and Josh Siegel,~\IEEEmembership{Senior Member,~IEEE}
\thanks{This paper was produced by the DeepTech Lab, within the Department of Computer Science and Engineering at Michigan State University.}
\thanks{Manuscript submitted to IEEE-IoT Journal September 19, 2024.}}

\markboth{IEEE Internet of Things Journal,~Vol.~X, No.~Y, Month~202X}%
{Shell \MakeLowercase{\textit{et al.}}: A Sample Article Using IEEEtran.cls for IEEE Journals}


\maketitle

\begin{abstract}
This paper introduces the Pareto Data Framework, an approach for identifying and selecting the Minimum Viable Data (MVD) required for enabling machine learning applications on constrained platforms such as embedded systems, mobile devices, and Internet of Things (IoT) devices. We demonstrate that strategic data reduction can maintain high performance while significantly reducing bandwidth, energy, computation, and storage costs. The framework identifies Minimum Viable Data (MVD) to optimize efficiency across resource-constrained environments without sacrificing performance. It addresses common inefficient practices in an IoT application such as overprovisioning of sensors and overprecision, and oversampling of signals, proposing scalable solutions for optimal sensor selection, signal extraction and transmission, and data representation. An experimental methodology demonstrates effective acoustic data characterization after downsampling, quantization, and truncation to simulate reduced-fidelity sensors and network and storage constraints; results shows that performance can be maintained up to 95\% with sample rates reduced by 75\% and bit depths and clip length reduced by 50\% which translates into substantial cost and resource reduction. These findings have implications on the design and development of constrained systems. The paper also discusses broader implications of the framework, including the potential to democratize advanced AI technologies across IoT applications and sectors such as agriculture, transportation, and manufacturing to improve access and multiply the benefits of data-driven insights. 

\end{abstract}

\begin{IEEEkeywords}
Machine Learning Efficiency, Minimum Viable Data, Resource-Constrained Computing, IoT Applications, Data Reduction Framework, Sensor Optimization, Sustainable Computing, Pareto Principle, Embedded Systems, Time-Series Analysis, AI Accessibility
\end{IEEEkeywords}


\section{Introduction}
\IEEEPARstart{I}{n} the Internet of Things (IoT), embedded systems, and constrained computing, the notion that “more data equals better performance” \cite{8493126} is being questioned. While large, high-quality datasets may yield valuable insights in some cases, the costs of capturing, transmitting, and storing such data can outweigh the performance gains, particularly in resource-constrained systems where sensor cost, computational complexity, energy, and network congestion are critical factors \cite{8972389}. In both academia and industry, there is a prevailing belief that using anything less than the best available data leads to poor outcomes (i.e., “garbage in, garbage out”). Although this holds true for safety-critical applications \cite{ao2023building,rengasamy2021towards}, in other domains it leads to unnecessary expense and complexity, sometimes discouraging the use of beneficial instrumentation altogether. 

As constrained systems become more prevalent, the mantra of ``more data equals better [machine learning] results'' creates a bottleneck incapable of being addressed by conventional Big Data tools alone, due in large part to resource constraints at the sensing device itself, and due to the complexity of introducing networking and backend infrastructure.  Importantly, not all data are equally valuable for every application \cite{jesus2017survey}, and overcapture results in ``data swamps'' in which the volume of data eclipses actionable intelligence. This prompts a key question: is more data really necessary for effective AI models? 

The need for smart data handling demands a shift towards frameworks that prioritize data quality and relevance over sheer volume. This is particularly important in constrained IoT devices, where limitations in battery life, bandwidth, computational power, and storage space demand efficient data use. This paper introduces ‘Minimal Viable Data’ (MVD), a sustainable approach that reduces data to the minimum necessary to meet performance goals. MVD refers to optimizing device configuration, sensor selection, and signal sampling to achieve application-specific performance with minimal resource use. Deploying sensors that capture less data becomes a strategic avenue to simultaneously reduce data storage and transfer costs while enhancing the efficiency of machine learning models, with potential initial and ongoing economic cost implications. 

Previous studies \cite{6252794,6824774,556484,casaseca2015effect,mollyn2022samosa} suggest that performance can remain strong even with limited or low-quality data. We hypothesize that there are key inflection points where the relationship between data input and output quality changes significantly, and that there exists a set of MVD sufficient for achieving target performance with minimal resource use.

In this article, we introduce the Pareto Data Framework, which seeks to optimize the balance between data quality, quantity, and resource consumption. Inspired by the Pareto Principle, which asserts that 80\% of effects come from 20\% of causes, this framework emphasizes capturing the most impactful data while minimizing resource use. This approach is particularly beneficial in IoT and related systems, where it optimizes resource constraints like battery life, bandwidth, computation, and storage.

Pareto Data Framework extends and generalizes our concept of Minimum Viable Data (MVD) by focusing on identifying and capturing the essential subset of data required for achieving specific objectives. This mitigates the constraints faced by IoT and related devices by championing targeting the MVD necessary to optimize the balance between data quantity, quality, and machine learning efficacy to facilitate high-efficiency applications even in resource-constrained settings. 

This concept applies across common IoT Machine Learning (ML) paradigms: CloudML, MobileML, and TinyML. For CloudML, MVD can reduce network latency and bandwidth use. For MobileML, it improves local runtime efficiency, and for TinyML, it makes resource-constrained applications feasible by reducing data requirements.

To explore the relationship between data quality and output performance, in this article, we evaluate system design and MVD through the use of time-series data to inform classification algorithms. We consider MVD optimization as a proxy for initial sensor cost, energy and storage consumption, and bandwidth requirements. We run experiments to validate our hypotheses and demonstrate how the Pareto Data Framework can generalize the MVD concept across applications. By promoting a focused, efficient data collection strategy, this framework has the potential to improve resource efficiency and make AI more sustainable and accessible in IoT systems.

Specifically, we consider the problems of overcapture, oversampling, and overprecision structured time-series data prevalent in many IoT applications through the use of audio signal inputs. Such data, often oversampled at 48kHz, could be downsampled, as neighboring points are highly correlated. Similarly, reducing precision (bit depth) and clip length can minimize resource use without degrading performance. Determining the optimal sample rate, bit depth, and clip length for AI models remains an open challenge.

Instead of indiscriminate data accumulation, the thoughtful curation of MVD as identified using the Pareto Data Framework can encourage a focused and efficient approach for determining the optimal sample rate, bit depth, and clip length for deep learning models in various applications from IoT to mobile and embedded computing. We explore the motivation and development for this framework in the following sections, before explaining the experimental methodology used in development and preliminary results. 

\section{Prior Work}
Efficient data use and sensor optimization in resource-constrained machine learning systems have been extensively studied, with many efforts focused on data selection \cite{siegel2016engine}, sensor optimization, and network efficiency \cite{siegel2016data}. This section critically reviews prior work to highlight existing limitations and demonstrate the need for a holistic and more comprehensive solution suitable for varied application domains, which we address with the Pareto Data Framework.

The data \textit{subset selection} problem \textendash choosing the most informative data under system constraints \textendash is known to be NP-hard \cite{natarajan1995sparse}, leading to multi-objective optimization algorithms. For instance, POSS\cite{qian2015subset} reduces subset size while optimizing selection criteria, but its \(2ek^2n\) makesuitable for large k and datasets. The distributed version (DPOSS) \cite{qian2018distributed} scales better but degrades significantly in noisy environments, as do related models \cite{8765790,qian2016parallel,roostapour2022pareto}, which suffer from computational inefficiency or lack generalizability for large-scale applications.

\textit{Data compression} techniques have been developed to tackle specific challenges. For example, \cite{7149287} assigns importance scores to sensor anomalies but struggles with quasi-periodic signals like ECG data. \cite{10297424} identifies points where data statistical properties change significantly, allowing the system to split the dataset into homogenous segments, though it may not perform optimally for less predictable signals. Real-world implementations \cite{10286737,9732548}, face deployment and scalability issues. Dynamic environments, like those face in synchronizing drones\cite{10286737} and network variability \cite{9732548} introduce constraints. These studies highlight the need for a more robust, generalizable framework that can effectively work across domains for data compression. 

\textit{Sensor selection} optimization involving a utility function been identified as NP-hard\cite{bian2006utility}. Evolutionary algorithms \cite{lin2017many} decompose large-scale IoT problems but rely on computationally expensive methods. Energy-efficient frameworks like WuKong \cite{huang2014energy} focus on communication energy consumption but lack empirical validation. Similarly, sensor reduction methods for specialized applications, such as medical shoes \cite{xu2015energy} and dynamic wireless sensor reconfiguration \cite{helkey2018sensor}, are effective but fail to generalize across broader IoT contexts due to predefined scenarios or scalability issues. Similarly other methods have been used \cite{Ghosh2021Learning-Based,Anvaripour2020A,Zhang2021SensorSF,Saucan2020Information-Seeking,zhang2020study,zhang2022frequency,damuut2013mixed}

\textit{Sparse data representation} has also been a prominent solution for reducing data volume\cite{ju2024comprehensive,peng2021low,sejdic2018compressive,KOUGIOUMTZOGLOU2020103082,parkale2017application,craven2014compressed,mishra2020soft,zhang2015survey}. These methods acquire compressed data directly, taking advantage of sparsity that may not be consistent across sensors, signals, and applications, limiting generalizability.

Some studies more directly informed our experiments, e.g., \cite{6252794} explored the impact of lowering sampling rates on feature efficacy, by quantizing raw data from 32 bit to 8 bits and downsampling from 44.1 kHz to 5.5kHz. \cite{6824774} proposed a compressive video sampling framework that optimizes the sampling rate and bit-depth to enhance the rate-distortion performance of video coding in resource-constrained environments
\cite{casaseca2015effect} finds that even with periodic downsampling (down to 400 Hz) the system maintains a high level of accuracy, suggesting that efficient, low-power cough monitoring via smartphones is feasible. Similarly, \cite{mollyn2022samosa} reduced power consumption in microphones while maintaining high recognition accuracy with lower audio sampling rates. 

These efforts demonstrate that resource-efficient data handling can maintain ML performance, but they lack a broader framework to generalize insights across different sensing applications.

The Pareto Data Framework, introduced in this paper, builds on these prior works by offering a holistic approach to data efficiency. Unlike previous efforts that focus on optimizing specific elements (e.g., data capture, transmission, or storage), our framework optimizes the entire data lifecycle. We address the challenge of balancing data fidelity with resource constraints by identifying the Minimum Viable Data (MVD)—the minimal data needed to meet performance targets in operational settings (unlike the definition provided in \cite{van2018prototyping,mosqueira2022classification}, which defines MVD as the minimum data necessary to train an ML model for early-phase agile AI prototyping). This enables scalable, resource-efficient IoT implementations and overcomes the limitations of prior domain-specific or element-focused optimizations.

\section{The Pareto Data Framework}
The Pareto Data Framework is a novel approach to addressing the challenges of data overabundance, sensor overprovisioning, and the high resource costs associated with traditional sensing systems and machine learning paradigms. This framework redefines how data are collected, processed, and utilized, particularly in resource-constrained environments where bandwidth, computation, storage, and energy are limited.

At its core, the framework distills data down to its most informative components by identifying and leveraging what we term ``Minimum Viable Data'' (MVD). This concept reduces the volume and fidelity of data required to achieve desired outcomes, ensuring both efficiency and sustainability in machine learning applications while maintaining high performance.

\begin{figure}[ht]
 \includegraphics[width=\linewidth]{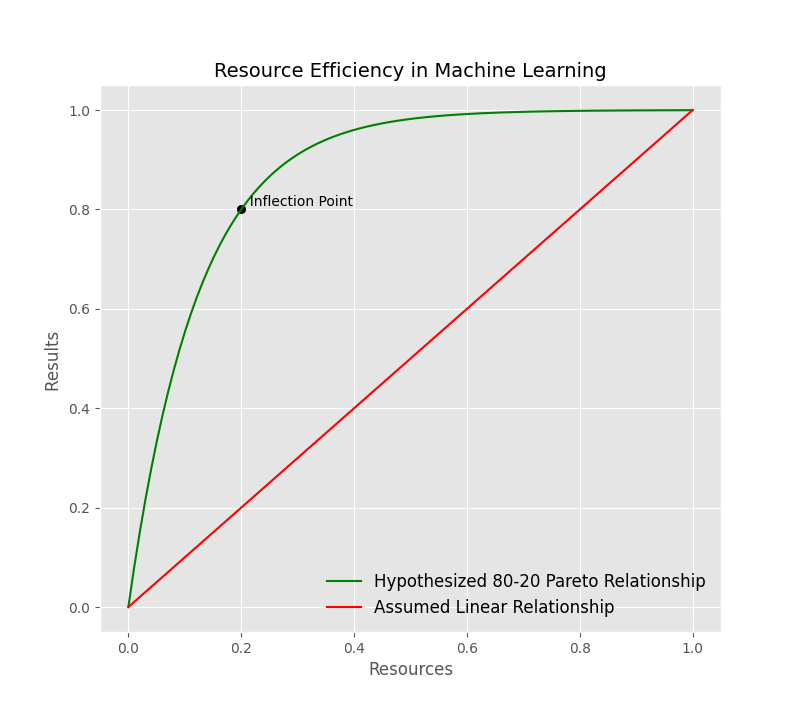}
 \caption{ Resources utilized up to the inflection point are considered the Minimum Viable Resources necessary for optimal results. Beyond this point, there is a cliff drop in performance gains. From a practical perspective, the implication are such that when assuming a linear relationship in a sensing system of a factory, a single high-quality sensor would be installed on one piece of equipment. However, by employing Minimum Viable Data (MVD) from the Pareto Data Framework, it becomes feasible to install 100 sensors across 100 pieces of equipment, offering a more comprehensive overview and data collection.
}
 
\label{fig:sample}
\end{figure}

The framework is driven by the hypothesis that there are common inflection points Fig \ref{fig:sample} in the data quantity-quality spectrum, beyond which additional data offers diminishing returns in relation to resource expenditure. By identifying these critical inflection points, the framework establishes MVD parameters that enable diverse applications to meet their performance targets with minimal resource use. This approach not only lowers costs and reduces the environmental footprint of data-intensive operations but also democratizes access to AI technologies, making them feasible in resource-constrained settings.

\subsection{Framework Objectives and Methodology}

The Pareto Data Framework is built around three objectives:

\begin{itemize}
 \item Resource Optimization: Minimize the resources required for data collection, transmission, and processing while maintaining acceptable levels of machine learning performance.
 \item Sustainability: Promote efficiency in AI and IoT deployments by reducing energy consumption, network congestion, and the need for high-end hardware.
 \item Accessibility: Reduce barriers to entry for implementing AI solutions in environments where economic, energetic, and other resource constraints might otherwise preclude their adoption. Similarly, increasing the scale of deployments for a fixed resource budget. 
\end{itemize}

To meet these objectives, the framework employs the following methodology:

\begin{itemize}
 \item Data Characterization: Qualitatively assessing data factors to identify those characteristics most critical to performance and those amenable to reduction or simplification without significant loss of utility.
 \item Experimental Validation: Conducting controlled experiments to empirically determine the MVD for various machine learning tasks, using metrics such as accuracy, precision, and recall as performance benchmarks.
 \item Generalization and Application: Extrapolating findings from specific case studies to broader applications, with a focus on generalizing the principles of MVD to a wide range of data types and machine learning paradigms.
\end{itemize}

\section{Experimental Setup}
We validate the concept of Pareto Data and Minimum Viable Data (MVD) through time-series audio data classification as a representative application domain. By experimenting with bit depth reduction, downsampling, and sample duration reduction, we simulate the resource constraints typical in sensor, bandwidth, processing, and power consumption scenarios and that impact both initial and operating cost. 

These experiments aim to generalize MVD through the Pareto Data Framework, offering validation for optimizing system design before sensor, network, and computing architecture deployment. By identifying inflection points in data quality and quantity, developers can predict performance outcomes and make informed resource allocation decisions. This approach ensures systems are optimized not only for performance but also for cost-efficiency, accessibility, and sustainability in long-term applications.



The objective is to identify permutable parameters relevant to resource utilization and to reduce them incrementally, mapping the performance degradation to pinpoint the “knee” or inflection point. This threshold reveals the minimum data quality and quantity needed before performance significantly declines, providing actionable insights for system optimization.


\subsection{Rationale for Audio Data Selection}
Audio data presents both challenges and opportunities of data optimization in resource-constrained settings \cite{10.1007/978-3-031-31327-1_12} and is ubiquitous in IoT applications ranging from smart home devices to urban noise monitoring systems. \cite{9099251} The inherent characteristics of audio signals, such as their temporal structure and the wide range of frequencies, make them an ideal candidate to investigate the effects of data reduction techniques like downsampling and quantization.  

Audio classification algorithms are sensitive to changes in sample rate, bit depth, and clip length. These properties make audio an ideal candidate for testing data reduction techniques that will scale to higher-dimensional signals. Further, the range of available microphones, from high-fidelity to lower-cost models, mirrors the quality-cost trade-offs faced in many real-world IoT deployments. Audio signals offer a robust framework for testing Pareto Data principles and allow us to generalize findings to other time-series and high-dimensional data domains while leveraging our extensive experience in acoustic characterization \cite{siegel2014vehicular,siegel2016data,siegel2015smartphone,terwilliger2022improving,siegel2021surveying}.

\subsection{Dataset Selection}
To ensure comprehensive evaluation of the Framework, we selected audio datasets that represent a range of real-world applications. Datasets with a higher number of classes reduce the likelihood of high performance due to random guessing. A larger class count also creates a greater margin of performance degradation as data quality and quantity are reduced. Additionally, datasets were selected to represent a range of data types, reflecting the variability encountered in different IoT applications. This diversity enables us to test the generalizability of sensor optimization techniques across various real-world scenarios. The chosen four datasets span different data types and class counts to maximize generalizability: (TABLE \ref{table1}):

The Environmental Sound Classification (ESC-50) dataset \cite{piczak2015esc} consists of 2,000 labeled audio across 50 classes of natural, human, and domestic sounds, offering a wide range of sound patterns and frequencies. It is ideal for evaluating how sensor optimization and data reduction techniques impact performance in diverse audio environments, such as environmental monitoring and smart cities, where noise and variability are prevalent. The GTZAN Music Genre Dataset \cite{tzanetakis_essl_cook_2001} contains 1,000 30-second of tracks across 10 music genres, recorded at 22,050Hz mono in 16-bit resolution. This dataset helps assess how reduced data quality affects classification accuracy in complex and overlapping sound patterns, often encountered in entertainment and audio recognition systems. The Toronto Emotional Speech Set (TESS) \cite{dupuis2010toronto} provides 1,400 speech recordings depicting seven emotional states, offering a valuable resource for studying subtle variations in human speech / audio pattern under constrained data conditions, crucial for applications like emotional recognition and virtual assistants. The Audio MNIST \cite{becker2024audiomnist} dataset features 30,000 spoken digit samples from 60 different speakers, making it ideal for evaluating the framework's performance when reducing sample rate, bit depth, and clip length in basic voice recognition tasks.

These datasets provide a comprehensive basis for testing and generalizing the concept and science behind the Pareto Data Framework.

\begin{table}[H]
\centering
\caption{Characteristics of Audio Datasets}
\renewcommand{\arraystretch}{1.5} 
\begin{tabular}{|m{1.5cm}|m{1.2cm}|m{1.2cm}|m{1.2cm}|m{1.2cm}|}
\hline
\centering \textbf{Dataset} & \centering \textbf{Sample Rate} & \centering \textbf{Bit Depth} & \centering \textbf{Clip Length} & \centering \textbf{Quantity} \arraybackslash \\ \hline
\centering ESC-50 \cite{piczak2015esc} & \centering 44,100 Hz & \centering 16 bits & \centering 5 sec & \centering 2,000 \arraybackslash \\ \hline
\centering GTZAN \cite{tzanetakis_essl_cook_2001} & \centering 22,050 Hz & \centering 16 bits & \centering 30 sec & \centering 1,000 \arraybackslash \\ \hline
\centering TESS\cite{dupuis2010toronto} & \centering 24,414 Hz & \centering 16 bits & \centering 1.5 sec & \centering 1,400 \arraybackslash \\ \hline
\centering Audio MNIST \cite{becker2024audiomnist} & \centering 48,000 Hz & \centering 16 bits & \centering 0.7 sec & \centering 30,000 \arraybackslash \\ \hline
\end{tabular}
\vspace{0.5cm}
\label{table1}
\end{table}

\subsection{Algorithms}
To evaluate the Pareto Data Framework across different machine learning models, we employed a variety of algorithms that balance resource efficiency and predictive performance. 

Decision trees are well-known for their simplicity and interpretability, and they are versatile in handling both categorical and numerical data \cite{osti_1839906}. The computational efficiency of decision trees is particularly beneficial in scenarios with reduced data, making them an attractive choice for resource-constrained environments \cite{10.1007/978-3-031-42536-3_10}. Extending this approach, the ensemble learning method Random Forest, built upon decision trees, presents an avenue for enhanced predictive performance \cite{DOMB2017218}. Logistic Regression, a powerful yet computationally efficient algorithm, serves as a valuable baseline model, especially in resource-light scenarios {\cite{9469605}} or where interpretability is paramount\cite{bhavitha2022novel} The simplicity and adaptability of K-Nearest Neighbors (KNN) make it well-suited for resource-limited scenarios\cite{8553155}, and has been effectively applied in analyzing data for predictive maintenance or for anomaly detection \cite{6879441}. Gradient Boosting models, exemplified by XGBoost or LightGBM, stand out as powerful ensemble methods capable of achieving high predictive performance even in the face of resource constraints \cite{Bentejac2021}. Additionally, certain neural network architectures, such as lightweight convolutional neural networks (CNNs) and shallow feedforward networks like MobileNet and SqueezeNet, are tailored for resource efficiency, catering to the demands of edge and mobile devices \cite{Yuan2022} . Support Vector Machines (SVMs), recognized for their effectiveness in binary and multiclass classification tasks, excel in managing reduced data facets and efficiently handling high-dimensional spaces \cite{Awad2015}\cite{Blanco2020}, thereby striking a fair balance between predictive performance and resource usage. 

Testing with these algorithms will help to demonstrate the framework's flexibility and applicability across different machine learning paradigms, establishing it's utility, relevance and applicability in real-world, resource-constrained environments. 

\subsection{Methodology}

\begin{figure*}[ht]
\centering 
\includegraphics[width=2.0
\columnwidth]{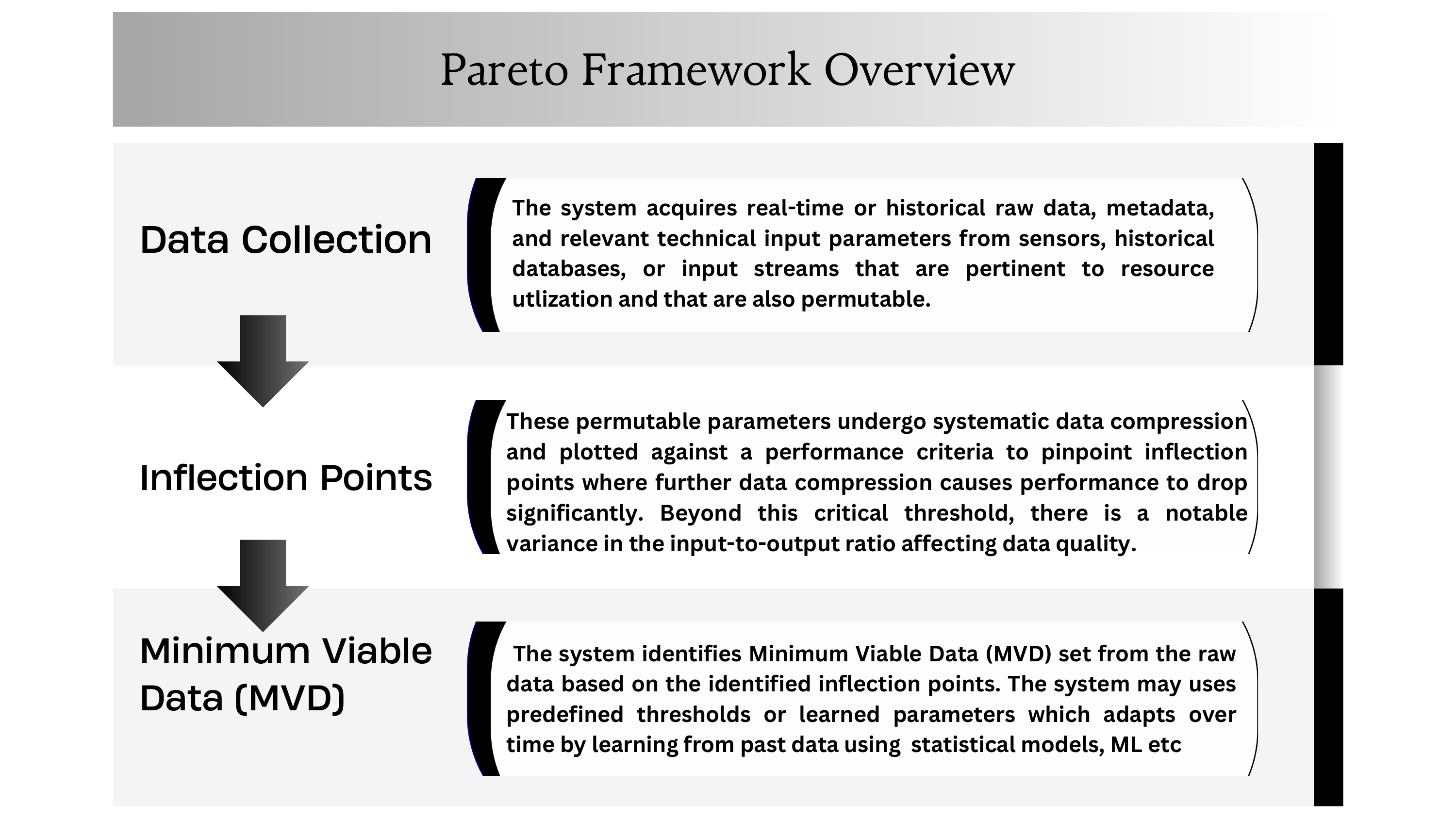}
\caption{An overview of the Pareto Data Framework methodology from raw-data to the MVD set.}
\label{fig:overview}
\end{figure*}

To optimize for resource constraints such as bandwidth, storage, sensor quality and cost, and computation in constrained systems, it is crucial to identify those data parameters that directly and significantly impact resources and that are quantifiable with the potential for alteration or reduction. 

For audio data, sample rate, bit depth, and clip length were selected as being significant as these have clear, measurable effects on resource consumption. Lower sample rates reduce the frequency range captured, leading to smaller file sizes and lower bandwidth requirements, while bit depth affects the precision of each sample, directly correlating with data rates and storage. Clip length impacts the amount of data retained for classification, and optimizing it can further reduce resource consumption. Other factors, such as the number of channels or coding formats, were considered less variable for reduction in this context.

The experiments were conducted in four phases, each focusing on a specific aspect of resource optimization:

\begin{enumerate}
 \item Downsampling: Sample rate determines the frequency range that can be accurately captured. Lower sample rates result in a reduced reliable frequency range, but offer smaller file sizes, lower bandwidth requirements, and potentially reduced energetic and economic costs as fewer data points are recorded per second. By adjusting the sample rates across a spectrum, this phase evaluated classification accuracy and measured benefits and tradeoffs of decreased data transmission. Sample rates of 44kHz, 22kHz, 16kHz, 8kHz, and 4kHz were considered.

 \item Quantization: Bit depth determines the precision of each sample in an audio. Lower bit depths directly correlate with reduced data rates and lower bandwidth consumption. Halving the bit depth from 16 bits to 8 bits also halves the data rate. The phase measured bandwidth savings weighed against the loss in classification accuracy. Each audio file was quantized to 16, 12, 10, 8, and 4 bit depth.
 
 \item Combined Sample Rate and Quantization: This phase simultaneously varied both sample rate and bit depth to analyze their combined impact on classification performance. By adjusting these parameters together, the aim was to better understand their interaction and potential benefits or detriments to performance and resource use. 

 \item Segmentation: This phase reduced clip length with the aim of ascertaining the minimum viable clip length that retains sufficient informational content for accurate classification, reflecting on storage and computational resource savings. The TESS and MNIST datasets began with 2 and 1 second samples, so had little opportunity for meaningful truncation. ESC\_50 began at 5 seconds and was segmented into clips from 1.0 to 5.0 seconds at half-second intervals. Similarly, the GTZAN dataset, with its original 30-second clips, was more gradually segmented down to intervals between 1 and 30 seconds, stepping down in increments of 10 seconds, 5 seconds, and finally 1 second.
\end{enumerate}

We selected specific values for periodic downsampling—sample rates of 44,100 Hz, 22,050 Hz, 16,000 Hz, 8,000 Hz, and 4,000 Hz—to represent a range from high-fidelity sensors (44,100 Hz), which are expensive and resource-intensive, to low-cost, energy-efficient sensors (4,000 Hz) suitable for resource-constrained environments; these values reflect  various sensor capabilities. For quantization, we chose bit depths of 16 bits, 12 bits, 10 bits, 8 bits, and 4 bits to simulate the trade-off between data precision and resource consumption: 16 bits represents high-precision sensors with greater costs and power needs, while 4 bits reflects minimal precision for ultra-low-cost sensors with minimal storage and energy requirements. Clip lengths were varied from 1 to 30 seconds (for the GTZAN dataset) and 1 to 5 seconds (for the ESC-50 dataset) to assess the impact on data volume and processing time, with longer clips representing detailed monitoring at higher operational costs, and shorter clips simulating more efficient data collection with reduced storage and computational demands. By experimenting with these specific values, we emulate a spectrum of sensor types and configurations that directly affect initial investments and ongoing operational costs

Prior to experimentation, each audio file underwent preprocessing, including normalization and feature extraction using Mel-frequency Cepstral Coefficients (MFCCs). MFCCs are commonly used in audio processing due to their effectiveness in representing the spectral properties of sound \cite{Sidhu2024}. For each audio file, MFCCs were computed using a 40-coefficient representation; these were the only features generated as they are effective at capturing the essential spectral characteristics of audio signals, particularly for resource constrained tasks.\cite{dey2020low,lan2014resource,ahmad2021audio} MFCCs reduce dimensionality and preprocessing complexity while maintaining moderate information fidelity and noise robustness. 

In each phase, to emulate the decision-making process that might be carried out by a sensor's data processing system in real-world applications, we chose SVM  classifier and accuracy as performance metric. It is one of the Algorithms discussed in the previous section that aligns closely with the operational constraints and efficiency needs of  sensor systems, \cite{ha2020machine}. MFCCs have also been shown to perform well with SVM \cite {7570164,godino2005support}. 
The graph was plotted for accuracy against varying levels of sample rate, bit depth, and clip length for each dataset. The results were plotted to identify the knee or inflection point—where accuracy begins to degrade sharply after incremental data reductions. The knee represents the practical limit for data optimization (the location of MVD), highlighting the point at which further reductions result in significant performance loss. Identifying this threshold is critical for balancing resource efficiency with performance. In some cases, the knee is clearly defined, while in others, the performance degradation may be more gradual, making it challenging to pinpoint an exact inflection point.

These experiments hold the potential to validate the concept of MVD by systematically exploring the trade-offs between data reduction and accuracy. The results provide a robust foundation for generalizing the Pareto Data Framework across a wide range of IoT and constrained computing applications, paving the way for more efficient, sustainable, and accessible machine learning solutions in resource-constrained environments.

\section{Results and Analysis}
This section highlights the trade-offs between resource use and machine learning performance. We identify the inflection points for ``minimum viable data'' (MVD) in various classification applications and across datasets, focusing on how these findings generalize and impact real-world IoT device optimization. By systematically reducing bit depth, sample rate, and clip length, we demonstrate the potential to balance data efficiency and computational performance, establishing the validity of the Pareto Data Framework as a means of identifying the Minimum Viable Data in certain contexts. Specific results follow.

\subsection{Sample Rate Reduction}
Reducing sample rates allowed us to identify the point at which lower frequencies begin to limit classification accuracy. Across all datasets, clear inflection points emerged, indicating where performance stabilized despite further reductions in sample rate.

In the TESS dataset, accuracy increased up to around 11,000 Hz, with diminishing returns beyond that point. For GTZAN, accuracy plateaued around 20,000 Hz. The MNIST dataset showed peak performance at a lower rate (~10,000 Hz), reinforcing the Pareto principle, as this dataset required less data to achieve high accuracy. ESC-50 exhibited a knee around 7,000 Hz, where further sample rate increases yielded marginal benefits.

In each dataset, the majority of the performance is achieved with a relatively low sample rate. These results highlight the potential to significantly reduce sample rates—sometimes down to 25\% of the original rate—while retaining 90-99\% of performance. This reduction translates into substantial bandwidth, energy, computation, and storage savings, reinforcing the applicability of the Pareto Data Framework to a wide range of IoT applications.


\begin{figure}[!t]
 \centering
 \begin{minipage}{0.50\linewidth}
  \centering
  \includegraphics[width=\linewidth]{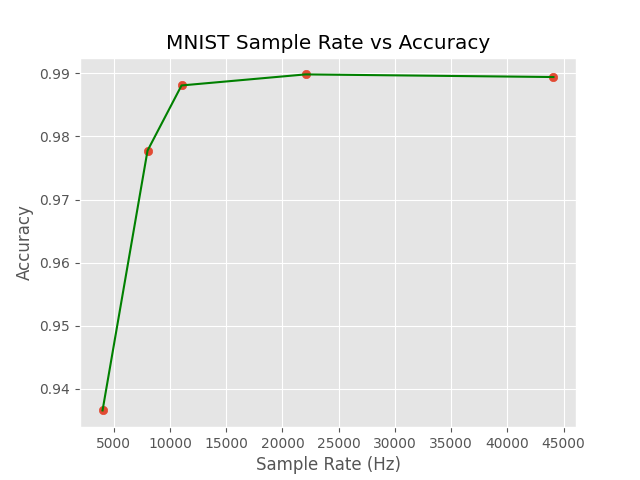}
  \caption*{MNIST}
  \label{srTESS}
 \end{minipage}\hfill
 \begin{minipage}{0.50\linewidth}
  \centering
  \includegraphics[width=\linewidth]{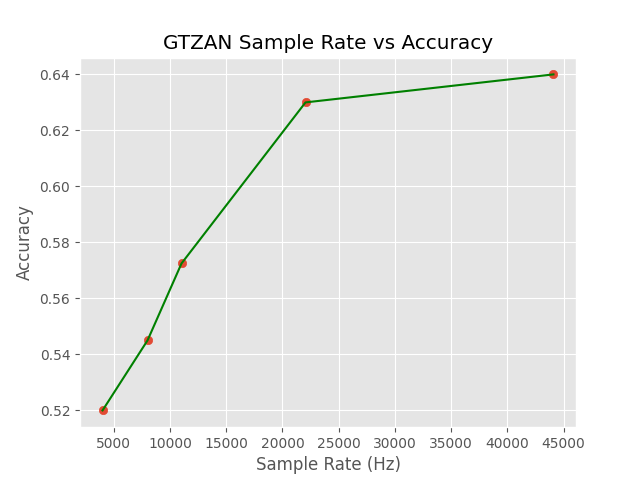}
  \caption*{GTZAN}
  \label{srGTZAN}
 \end{minipage}

 \begin{minipage}{0.50\linewidth}
  \centering
  \includegraphics[width=\linewidth]{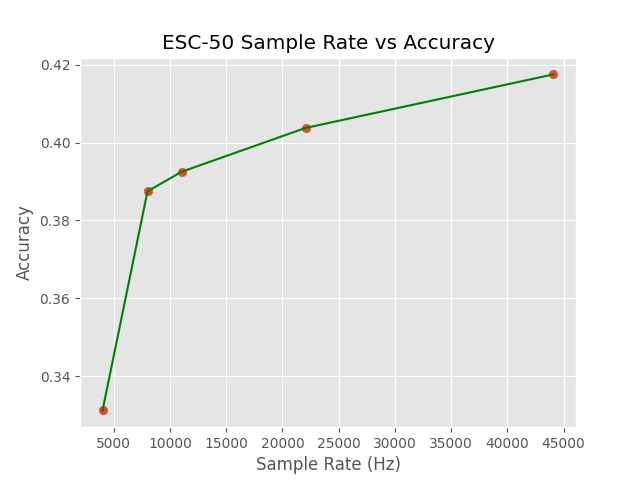}
  \caption*{ESC-50}
  \label{srESC}
 \end{minipage}\hfill
 \begin{minipage}{0.50\linewidth}
  \centering
  \includegraphics[width=\linewidth]{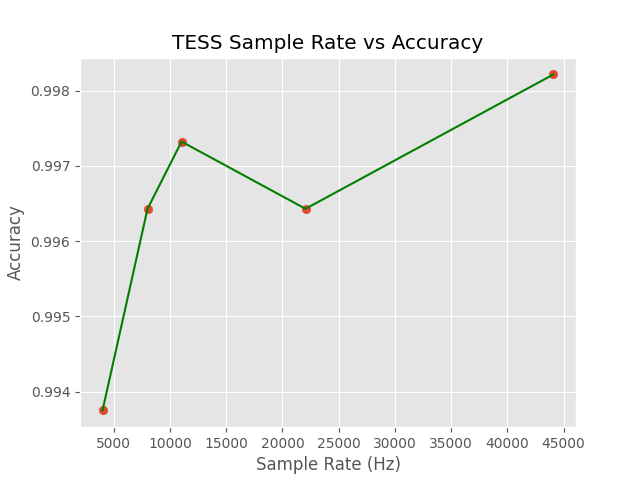}
  \caption*{TESS}
  \label{srMNIST}
 \end{minipage}
 
 \caption{Impact of Sample Reduction on Classification Accuracy}
 \label{fig:sr}
\end{figure}

\subsection{Bit Depth Reduction}

\begin{figure}[!t]
 \centering
 \begin{minipage}{0.50\linewidth}
  \centering
  \includegraphics[width=\linewidth]{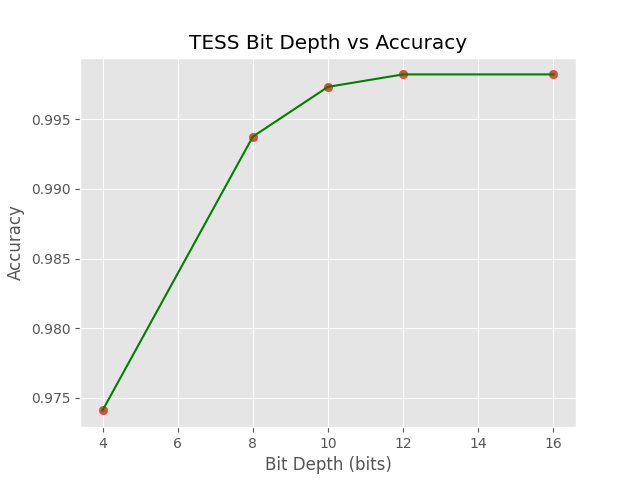}
  \caption*{TESS}
  \label{fig:bitdepth_reduction1}
 \end{minipage}\hfill
 \begin{minipage}{0.50\linewidth}
  \centering
  \includegraphics[width=\linewidth]{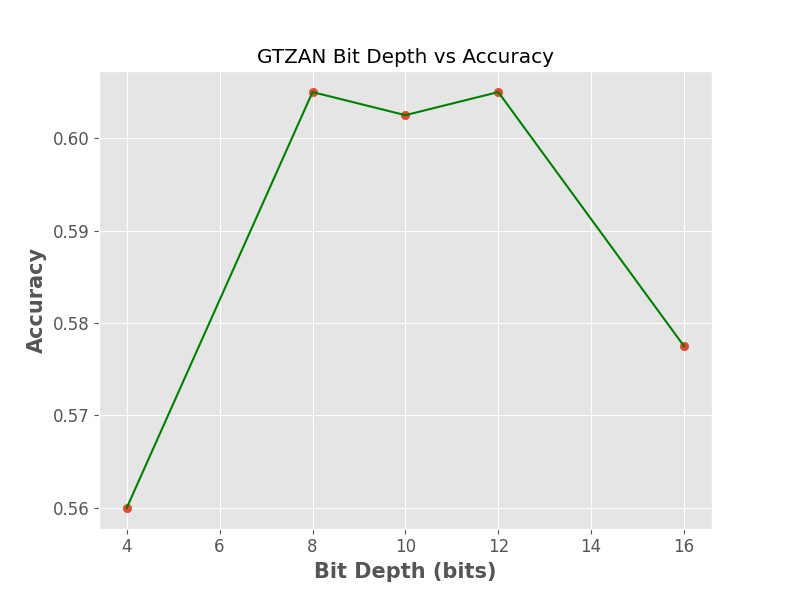}
  \caption*{GTZAN}
  \label{fig:bitdepth_reduction2}
 \end{minipage}

 \vspace{0.5cm}

 \begin{minipage}{0.50\linewidth}
  \centering
  \includegraphics[width=\linewidth]{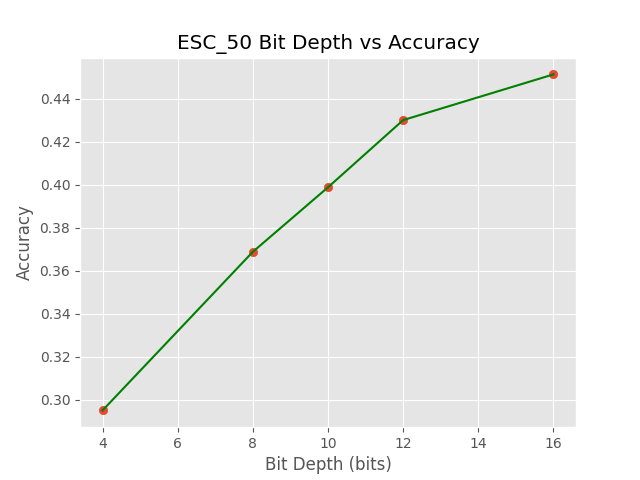}
  \caption*{ESC-50}
  \label{fig:bitdepth_reduction3}
 \end{minipage}\hfill
 \begin{minipage}{0.50\linewidth}
  \centering
  \includegraphics[width=\linewidth]{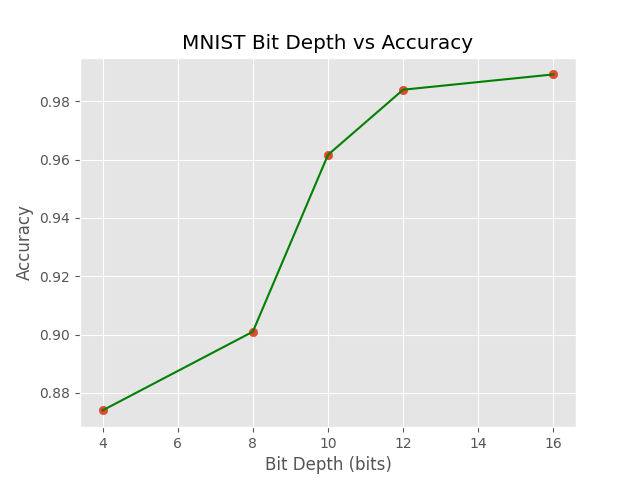}
  \caption*{MNIST}
  \label{fig:bitdepth_reduction4}
 \end{minipage}
 
 \caption{Impact of Bit Depth Reduction on Classification Accuracy}
 \label{fig:bit_depth}
\end{figure}

Bit depth reduction explored the impact of quantization from 4- to 16-bit depth on performance across datasets, with a focus on balancing precision and resource usage (bit depth as a proxy for economic, energetic, and network costs). The results revealed inflection points where further reductions in bit depth began to degrade classification accuracy.

For the TESS dataset, accuracy increased sharply as bit depth rose from 4 to 8 bits and stabilized after 10 bits, indicating that 8-10 bits is the most resource-efficient range. MNIST followed a similar trend, with the optimal performance occurring between 10-12 bits. For GTZAN, the knee point occurred at 8 bits, with further increases introducing minimal gains or slight decreases, likely due to the complexity and excessive granularity of the audio patterns as the precision increases, causing an overfit. ESC-50 showed a consistent increase in accuracy as bit depth rose, with no clear knee point, suggesting a more linear relationship between bit depth and performance in this dataset, suggesting fewer potential ``savings'' in quantization, and indicating that differing classification tasks may offer varied knee locations, or linearity of input to output quality.

We note that in ESC\_50, bit depth does make an impact in a later section when reduced simultaneously with sample rate, showing that some parameters of data might not make impact individually but still show potential for a characteristic inflection point. 

As expected, higher bit depths initially exhibited higher accuracy, capturing finer details and nuances in the audio data. A common finding was that accuracy can be largely maintained even with significant reductions in bit depth. One plausible reason for this is that beyond a certain threshold, increasing bit depth captures non-informative features, including additional noise. 

The results support the MVD principle by showing how data collection can be optimized for performance without unnecessary resource consumption. Reducing bit depth by half cuts bandwidth by 50\%, demonstrating the potential for significant resource savings without sacrificing accuracy.


\subsection{Combination of bit Depth and Sample rate Reduction}
\begin{figure}[!t]
 \centering
 \begin{minipage}{0.45\linewidth}
  \centering
  \includegraphics[width=\linewidth]{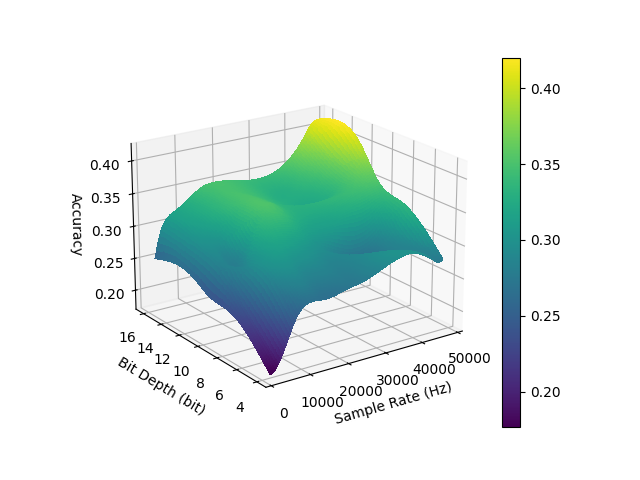}
  \caption*{ESC-50}
  \label{bothESC}
 \end{minipage}\hfill
 \begin{minipage}{0.45\linewidth}
  \centering
  \includegraphics[width=\linewidth]{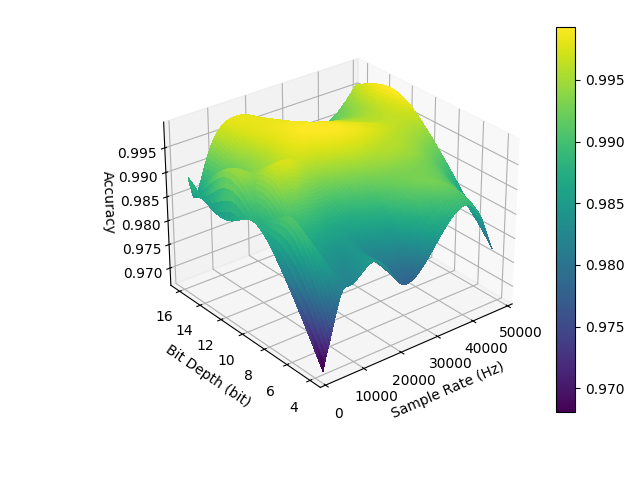}
  \caption*{TESS}
  \label{bothTESS}
 \end{minipage}

 \begin{minipage}{0.45\linewidth}
  \centering
  \includegraphics[width=\linewidth]{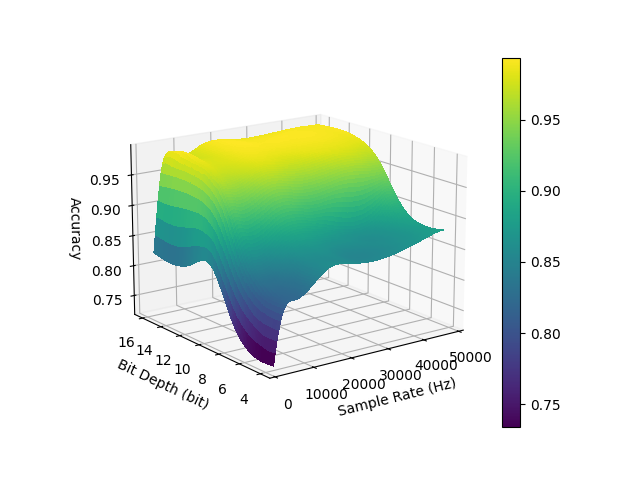}
  \caption*{MNIST}
  \label{bothMNIST}
 \end{minipage}\hfill
 \begin{minipage}{0.45\linewidth}
  \centering
  \includegraphics[width=\linewidth]{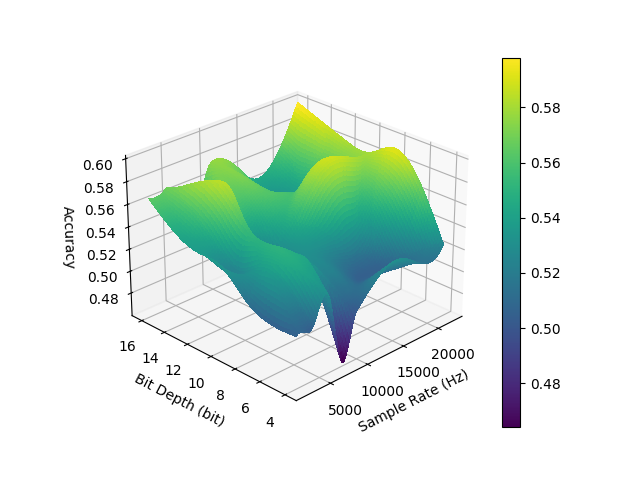}
  \caption*{GTZAN}
  \label{bothGTZAN}
 \end{minipage}
 
 \caption{Impact of Simultaneous Bit Depth-Sample rate Reduction on Classification Accuracy}
 \label{fig:bitdepth_reduction}
\end{figure}

When bit depth and sample rate were reduced simultaneously, the results further validated the Pareto Data Framework. The TESS and ESC-50 datasets showed accuracy stabilizing after key thresholds (20,000 Hz and 8 bits for TESS, and 20,000 Hz and 12 bits for ESC-50). GTZAN displayed more fluctuation, but the general trend confirmed that most accuracy gains could be achieved with moderate levels of both sample rate and bit depth.

For MNIST, the knee occurred earlier, with high accuracy retained even at lower data quality levels. This emphasizes that simpler datasets benefit from more aggressive data reduction, making them prime candidates for resource-efficient implementation in constrained environments.

The combined approach confirms that both dimensions can be reduced in tandem, offering a multi-dimensional method for optimizing data efficiency without sacrificing performance.

This multidimensional reduction reasoning allows for a more holistic optimization process, where multiple data dimensions are fine-tuned in tandem to reach the collective inflection point demonstrating the Framework's applicability across diverse and complex scenarios.

\subsection{Clip Length}
Shortening audio clip length helped to identify the minimum viable duration for maintaining classification accuracy.  For GTZAN (with original clip length of 30s), accuracy improved as chunk size increased up to 15 seconds, after which performance plateaued. Similarly, ESC-50 saw performance stabilize at 2.5 seconds, suggesting that clip lengths could be reduced while maintaining up to 95\% of the original accuracy.

\begin{figure}[!t]
 \centering
 \begin{minipage}{0.50\linewidth}
  \centering
  \includegraphics[width=\linewidth]{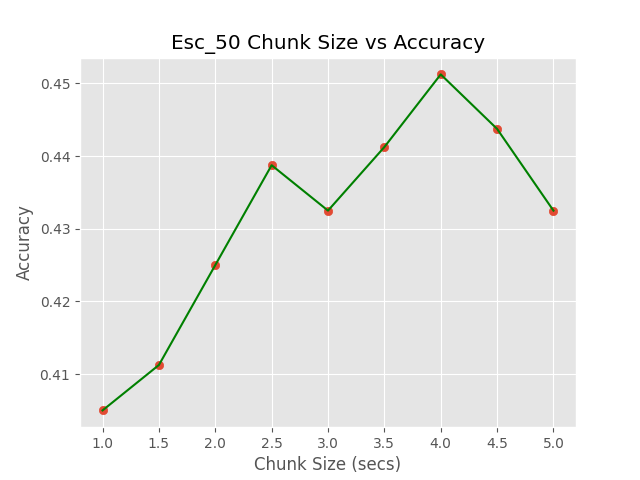}
  \caption*{ESC-50}  
  \label{chunkESC}
 \end{minipage}\hfill
 \begin{minipage}{0.50\linewidth}
  \centering
  \includegraphics[width=\linewidth]{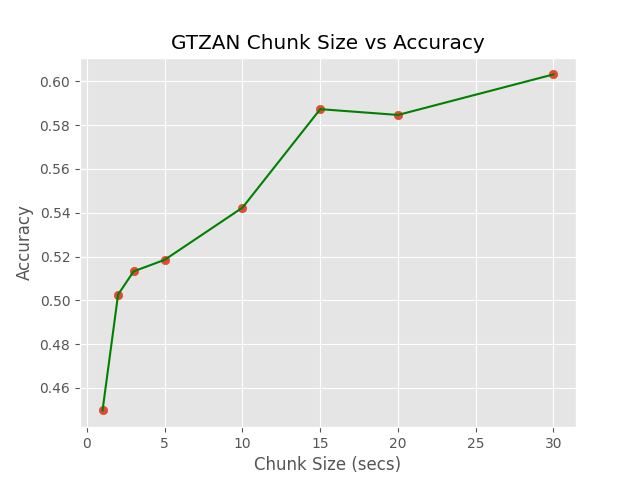}
  \caption*{GTZAN}  
  \label{gtzan_chunk}
 \end{minipage}

 \vspace{0.5cm}

 \caption {Periodic Clip Length  Reduction and its impact on Classification Accuracy}  
 \label{fig:clip_length}
\end{figure}

Reduction through this framework could potentially halve the file size (File size = Bitrate x Clip Length), while still retaining performance of up-to 95\% of original accuracy, (as shown in Fig \ref{fig:clip_length}) leading to more efficient storage and processing. 

By reducing clip lengths, we achieved substantial reductions in file size, improving storage and processing efficiency without major losses in performance. This further validates the flexibility of the Pareto Data Framework in adapting to various data characteristics and application needs.

\subsection{Generalizability}
While the above results are demonstrated on audio data, the principles of data reduction through MVD can generalize to other time-series data such as sensor networks, visual data streams, and environmental monitoring. In these domains, parameters such as frame rate (for video) or sampling intervals (for sensor networks) can be optimized similarly to achieve resource-efficient machine learning.

In real-world, for a given application particularly when no prior collected dataset is available, a practical approach is to tune the application parameters incrementally adjusting them until a noticeable dip in performance occurs, pinpointing this inflection point as the threshold for optimal data use. In context of broader deployment of this framework, which involves dealing with several related applications to those previously studied, the established inflection points provide a useful starting guideline, allowing for minor adjustments based on specific data characteristics. It can also involve utilizing statistical and learning models over the previously studied relationship between data quality and model performance, thereby reducing the need for extensive experimentation.
ns such as environmental monitoring, smart cities, and industrial IoT, where efficient data collection and processing are critical.

For instance, in smart city infrastructure, the framework offers a means to optimize data collection from sensors monitoring traffic, pollution, and public safety. By reducing energy, bandwidth, and operational costs, the framework can enable more efficient real-time decision-making without overwhelming the underlying IoT infrastructure. Similarly, in precision agriculture, the framework allows for the strategic deployment of sensors to monitor soil conditions, equipment health, and environmental factors, making these advanced technologies more accessible to small and medium-sized farms. This scalability enhances precision agriculture practices by making sophisticated monitoring tools more cost-effective.

The framework also has significant implications for transportation and infrastructure. In off-board vehicle monitoring, for example, using less sensitive microphones and lower sampling rates still allows for the accurate detection of anomalies, such as engine knocks or exhaust leaks. This enables manufacturers to deploy diagnostic systems more widely across fleets without increasing costs or resource consumption. The broader adoption of acoustic diagnostics~\cite{siegel2014vehicular,siegel2015smartphone,siegel2016engine,siegel2016data,siegel2017air,siegel2018automotive, siegel2021surveying} can improve vehicle safety and reliability, as well as reduce maintenance costs.

\subsection{Factory Example}
Consider a factory manager tasked with setting up and operating a sensing system for 10 years with a budget of \$1000. In a traditional approach, with an assumed linear relationship between cost and data quality, the manager could install a single high-quality sensor on one piece of equipment. This sensor would provide detailed, lab-grade data, but it would only offer insight into that single machine—leaving the rest of the factory unmonitored.

With the Pareto Data Framework and the concept of Minimum Viable Data, the same budget can instead be used to install 100 lower-cost vibration  
 \cite{ siegel2017future,siegel2016data, 8263445, Siegel2020} or current~\cite{siegel2018real} sensors across 100 pieces of equipment. While each sensor may not capture the highest possible fidelity, together they provide a comprehensive, system-wide view of the factory's operations—the ``30,000 foot view.'' These sensors deliver directional insight, enabling the manager to observe general trends in performance, detect inefficiencies, and anticipate maintenance needs across the entire operation.

This broader sensor network can serve as the foundation for an effective decision support system. By collecting data from multiple points, the system can identify patterns, relationships, and trends that would be invisible with isolated high-fidelity data. For instance, a drop in power consumption across several machines might indicate the onset of wear in a specific part of the production line, prompting proactive maintenance before a breakdown occurs. Similarly, real-time insights from many sensors allow the factory manager to adjust production schedules dynamically, optimize resource allocation, and improve overall throughput.

Even though the data from each sensor is not perfect, its aggregation allows for a deeper understanding of the factory's operations, creating a feedback loop that continuously informs decision-making. The ability to monitor more equipment and track long-term performance metrics leads to better forecasting of maintenance needs, reducing unexpected downtime and extending the lifespan of critical machinery. Additionally, the cumulative insight from this distributed network supports more informed decisions about energy consumption, quality control, and efficiency improvements, which can drive significant cost savings over time.

In this way, the Pareto Data Framework transforms a limited budget into a scalable, intelligent monitoring solution. It maximizes the utility of available resources by emphasizing breadth of coverage rather than precision at a single point, helping the factory manager make smarter, data-driven decisions that enhance operational efficiency and support the factory's long-term growth.

\section{Conclusion and Future Work}
In this paper, we introduced the Pareto Data Framework, a novel approach for optimizing data collection and processing in resource-constrained environments. Our experiments with sensor data proxies confirm that optimizing parameters like sample rate, bit depth, and clip length significantly reduces resource consumption while preserving performance, validating the Pareto Data Framework. Future work will extend the framework to domains such as visual and environmental data, further refining MVD for broader IoT applications. Our results validated the presence of inflection points in the data, where the relationship between input and output quality changes significantly. This finding challenges the prevailing notion that more or higher-quality data always leads to better outcomes. The framework paves the way for more sustainable AI and machine learning practices, enabling broader deployment of these technologies across industries and regions where resource constraints have been barriers to progress.

The Pareto Data Framework offers immediate practical benefits for a wide range of IoT applications. By focusing on capturing only the most essential data, systems can be designed to operate efficiently within resource limits. In industrial IoT, for instance, this approach enables the deployment of more sensors across more machines, providing broader operational insights without incurring excessive costs. In mobile and edge AI, reducing data fidelity can extend battery life and lower energy consumption, making advanced AI capabilities feasible in resource-limited settings. Similarly, in environmental monitoring, lower power requirements and smaller data transmissions allow for longer-term, remote deployment of sensors.

However, the exploration of the Pareto Data Framework is only beginning. Future research will extend its application beyond acoustic data to other domains such as acceleration, visual data, and complex sensor networks. Each of these domains presents unique challenges that will require the development of new methodologies and algorithms tailored to their specific data characteristics. This will further refine the concept of Minimum Viable Data (MVD), maximizing data efficiency while preserving the utility of machine learning insights.

Additional exploration into the integration of the Pareto Data Framework with edge computing offers promising opportunities. By bringing data processing closer to the source, edge computing complements data efficiency, potentially enhancing the framework’s application in real-time, latency-sensitive use cases.

Our long-term goal is to generalize the framework’s principles, developing mathematical models that can predict optimal data reduction strategies across various applications, with or without representative data. This effort will distill the insights from our empirical studies into a set of guiding principles and tools for practitioners. Future work will also focus on more rigorous evaluation methods that account for total resource costs in specific operating contexts, ensuring that the Pareto Data Framework continues to drive innovation in resource-efficient AI deployment.

\section{Funding}
This work was supported under the MTRAC Program by the State of Michigan 21st Century Jobs Fund received through the Michigan Strategic Fund and administered by the Michigan Economic Development Corporation.

\bibliographystyle{IEEEtran}
\bibliography{references} 

\vfill

\end{document}